\documentclass[english,letterpaper, 10pt, conference]{icra09}
\usepackage[T1]{fontenc}
\usepackage[latin9]{inputenc}
\usepackage{array}
\usepackage{rotating}
\usepackage{graphicx}

\makeatletter

\providecommand{\tabularnewline}{\\}
\newcommand{\lyxdot}{.}

\IEEEoverridecommandlockouts
\usepackage[footnotesize]{caption}
\usepackage{cite}

\@ifundefined{showcaptionsetup}{}{%
 \PassOptionsToPackage{caption=false}{subfig}}
\usepackage{subfig}
\makeatother

\usepackage{babel}
\begin{document}

\title{\LARGE \bf Utilizing Semantic Visual Landmarks for Precise Vehicle
Navigation}

\author{Varun Murali, Han-Pang Chiu,\\
Supun Samarasekera, Rakesh (Teddy) Kumar\thanks{The authors are with Center for Vision Technologies, SRI International, Princeton, NJ 08540, USA. The contact authors are Varun Murali {\tt \small \{varun.murali@sri.com)} and Han-Pang Chiu {\tt \small \{han-pang.chiu@sri.com)}} }
\maketitle
\begin{abstract}
This paper presents a new approach for integrating semantic information
for vision-based vehicle navigation. Although vision-based vehicle
navigation systems using pre-mapped visual landmarks are capable of
achieving submeter level accuracy in large-scale urban environment,
a typical error source in this type of systems comes from the presence
of visual landmarks or features from temporal objects in the environment,
such as cars and pedestrians. We propose a gated factor graph framework
to use semantic information associated with visual features to make
decisions on outlier/ inlier computation from three perspectives:
the feature tracking process, the geo-referenced map building process,
and the navigation system using pre-mapped landmarks. The class category
that the visual feature belongs to is extracted from a pre-trained
deep learning network trained for semantic segmentation. The feasibility
and generality of our approach is demonstrated by our implementations
on top of two vision-based navigation systems. Experimental evaluations
validate that the injection of semantic information associated with
visual landmarks using our approach achieves substantial improvements
in accuracy on GPS-denied navigation solutions for large-scale urban
scenarios. 
\end{abstract}

\section{Introduction}

Vehicle navigation using a pre-built map of visual landmarks has received
lots of attention in recent years for future driver assistance systems
or autonomous driving applications \cite{guizzo2011google}, which
require sub-meter or centimeter level accuracy for situations such
as obstacle avoidance or predictive emergency braking. The map of
the environment is constructed and geo-referenced beforehand, and
is used for global positioning during future navigation by matching
new feature observations from on-board perception sensors to this
map. Due to the low cost and small size of camera sensors, this approach
is more appealing than traditional solutions using costly and bulky
sensors such as differential GPS or laser scanners \cite{levinson2010robust}. 

\begin{figure}[t]
\centering\includegraphics[scale=0.28]{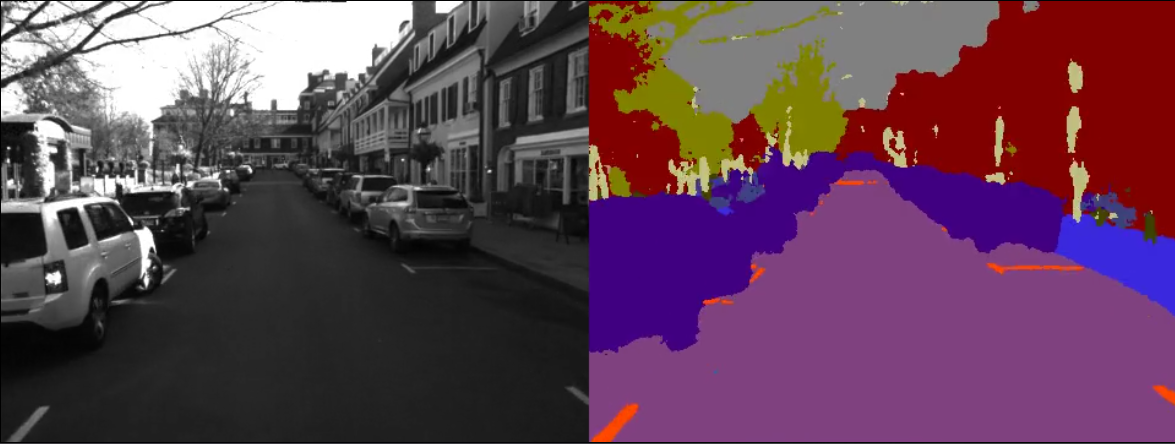}

\caption{\label{fig:sem_seg}Output of a pre-trained Semantic Segmentation
Network \cite{badrinarayanan2015segnet} used on one of our datasets.
The figure above shows a use case for where semantic information is
useful in mapping. Features from the parked vehicles can be considered
to be temporal and should not be maintained in a long term map.}
\end{figure}

Using visual information from permanent structures rather than temporal
objects ought to improve the mapping quality and navigation accuracy
for these vision-based navigation systems. With new advances in deep
learning, previously hard computer vision problems such as object
recognition and scene classification can be solved with high accuracy.
The availability of these trained models is able to make the use of
vison-based navigation algorithms easier. For example, Figure \ref{fig:sem_seg}
shows the result of using an off-the-shelf video segmentation tool
to classify object categories from a street scene. As can be seen
from the figure, even pre-trained networks achieve high accuracy and
can help the navigation problem.

A recent survey on Simultaneous Localization and Mapping (SLAM) \cite{cadena2016simultaneous}
shows the frontiers that have been explored in adding semantic information
to the mapping process. They discuss the use of semantic information
in mapping, and present arguments for both why semantic information
aids SLAM and vice versa. These works have used different approaches
to map the environment with different representations for semantic
information in the environment. For instance, Choudhary et al \cite{choudhary2014slam}
map indoor environments at the level of objects. Other works map the
environment at the level of planes. These works highlight the importance
of semantic mapping, which allows humans to localize themselves because
they are able to associate geometric features with semantic categories.

However, the map maintained in these systems only preserves high-level
objects/ planes or other semantic entities. These techniques are typically
used in the domain of mobile robots that operate indoors. They are
interested in maintaining representations of objects or locations
of obstacles (such as walls) that the robot can maneuver, and are
not directly applicable to the autonomous vehicle navigation problem
we try to solve. In our application, it is important to maintain both
high-level semantic information and low-level visual features associated
with landmarks mapped in the environment. In addition, these works
use complex algorithms to perform image/video segmentation to derive
semantic information for the localization and mapping process. With
recent advances in deep learning, segmentation tasks can be replaced
with simpler and off the shelf tools.

In this paper, we present a simple and effective approach for integrating
semantic information extracted from a pre-trained deep learning network
for vehicle navigation. We propose a new framework which utilizes
the semantic information associated with each imaged feature to make
a decision on whether to use this feature point in the system. This
feature selection mechanism based on semantic information can be performed
for both the feature tracking process in the real-time navigation
system and the map building process beforehand. For example, as shown
in Figure \ref{fig:sem_seg}, imaged features from the parked vehicles
can be considered to be temporal and should not be maintained in the
map during the map building process.

Our framework is designed to be applied to any vision-based SLAM systems
for vehicle navigation. To demonstrate the feasibility and generality
of our approach, we apply our framework on two state-of-the-art visual-based
navigation systems \cite{chiu2016navigation,ORBSLAM}. The popular
ORB-SLAM2 system \cite{ORBSLAM} uses only cameras as the input sensor,
and cannot build geo-referenced maps beforehand. Without the use of
pre-built maps, we show our approach improves its navigation accuracy
by polishing its feature tracking process. 

We also utilize the system from \cite{chiu2016navigation}, which
efficiently fuses pre-mapped visual landmarks as individual point
measurements to achieve sub-meter overall global navigation accuracy
in large-scale urban environments. This system constructs the visual
map beforehand by using a monocular camera, IMU, and high-precision
differential GPS. Our approach improves both the mapping quality and
the tracking process in this system, and achieves approximately 20\%
improvement in accuracy for its GPS-denied navigation solution.

The rest of the paper is organized as follows. In Section II, we present
the related work for vehicle navigation both with and without semantic
information. In Section III, we introduce a new gated factor graph
framework, which incorporates semantic information in the factor graph
formulation for vision-based vehicle navigation systems. We present
how our framework is used for improving the feature tracking process,
constructing the landmark database during the pre-mapping process,
and achieving high-precision navigation performance using pre-mapped
landmarks. In Section IV, we present our experimental setup and our
experimental results on top of two state-of-the-art vision-based navigation
systems. Finally the conclusions and future work is presented in Section
V. 

\section{Related Work}

Traditional solution to achieve high-level accuracy for vehicle navigation
is to fuse high precision differential GPS with high-end inertial
measurement units (IMUs), which is prohibitively expensive for commercial
purpose. Non-differential GPS can be cheap, but rarely reach satisfactory
accuracy due to signal obstructions or multipath effects.

Many methods for visual SLAM \cite{durrant2006simultaneous,fuentes2015visual,davison2007monoslam,davison2003real,ORBSLAM}
which solve the joint problem of state estimation and map building
simultaneously in unknown environments, have been proposed for vehicle
navigation applications. Visual inertial navigation has also been
extensively studied \cite{chiu2013robust,leutenegger2015keyframe}
for using feature tracks and IMU measurement to solve the navigation
problem. However, without absolute measurements, none of these methods
maintain overall sub-meter global accuracy within large-scale urban
environments.

There are a few methods which incorporate semantic information into
SLAM systems for vehicle navigation. For example, Reddy et al \cite{reddy2015dynamic}
use a multi-layer conditional random field (CRF) framework to perform
motion segmentation and object class labeling. It improves localization
performance by only mapping stationary objects in the environment
and excluding dynamic objects from the scene. The semantic motion
segmentation is computed from the disparity and computed optical flow
between inputs. Another method \cite{kundu2014joint} proposes to
use CRF and factor graph to jointly model the environment and recover
the pose of the navigation platform. They also use labels and visual
SLAM to densely reconstruct the 3D environment. 

Using a pre-optimized visual map is able to achieve high-level accuracy
for vehicle navigation, by matching new feature observations to mapped
landmarks. There are a number of methods \cite{chiu2016navigation,beall2014appearance}
that propose to use cameras to construct a geo-referenced map of visual
landmarks beforehand. Each optimized visual landmark in the map consists
its absolute 3D coordinate, with its 2D locations and visual descriptors
from 2D images. This map of 2D-3D visual landmarks then provides absolute
measurements for future vehicle navigation. However, none of these
works utilize semantic information to improve the mapping quality
or the navigation accuracy.

Recently, Alcantarilla et al \cite{alcantarilla2016street} present
a way to incorporate semantic information to improve the mapping quality.
They use a deconvolution network to classify changes between two query
images taken from the same pose in the world. They present a dataset
with annotation for changes in images taken from the same pose at
different times. They train a deconvolutional network on this data,
and show that a network can be learnt effectively to detect changes
in street view images. There are also approaches based on information
theory to reduce the number of landmarks in the visual codebook, such
as \cite{choudhary2015information}. They use information theoretic
heuristics to remove landmarks that are adding little value.

The closest work to our presented framework is the episodic localization
approach proposed by Biswas et al \cite{biswas2014episodic} using
a varying graphical network. They propose a method that distinguishes
temporal and permanent objects based on the reprojection error of
the feature. This allows them to classify the features into long and
short term features in the map. They also modify the cost function
at every timestep with the current estimate of the feature being long/
short term to improve navigation accuracy. Temporary maps \cite{meyer2010temporary}
is another approach that is proposed to keep track of newly appearing
objects in the environment or objects that have not been previously
mapped to improve navigation accuracy. All these works focus on indoor
navigation, and are not directly applicable to autonomous vehicle
navigation applications.

In contrast to previous works, our approach improves both the tracking
process and the mapping quality by selecting visual landmarks based
on semantic categories associated with the extracted features. Our
framework is also easy to be formulated, and can be applied in any
vision-based navigation systems. We show that our approach uses semantic
information to improve vehicle navigation performance, both with and
without the use of pre-mapped visual landmarks. 

Note the main interest of this work is the overall global navigation
accuracy including places where only few or no valid visual landmarks
are available due to scene occlusion or appearance change. Thus, we
evaluate our GPS-denied navigation accuracy against ground truths
provided by fusing high-precision differential GPS with IMUs, which
is different from calculating the localization error as the relative
distance to the visual map \cite{IV13}.

\section{Approach\label{sec:Approach}}

This section describes our approach to incorporate semantic information
for vision-based vehicle navigation. Note our approach is generic
to any vision-based SLAM systems for vehicle navigation. However,
for implementation and demonstration purposes, our approach is built
on top of two state-of-the-art vision-based navigation systems \cite{chiu2016navigation,ORBSLAM}. 

Our approach improves the system performance in three ways: the feature
tracking process, the map building process, and the navigation accuracy
using pre-mapped landmarks. The ORB-SLAM2 system \cite{ORBSLAM} has
been proposed for vehicle navigation applications without the use
of GPS or geo-referenced visual maps. For this system, our approach
utilizes the semantic information to improve its feature tracking
process during navigation.

The tightly-coupled visual-inertial navigation system in \cite{chiu2016navigation}
efficiently utilizes pre-mapped visual landmarks to achieve sub-meter
overall global accuracy in large-scale urban environments, using only
IMU and a monocular camera. It also builds a high-quality, fully-optimized
map of visual landmarks beforehand using IMU, GPS, and one monocular
camera. Our approach incorporates semantic information in this system
for both the map building process and GPS-denied navigation using
pre-mapped visual landmarks. 

\subsection{Semantic Segmentation\label{subsec:Semantic-Segmentation}}

The semantic segmentation for the input sequence is processed using
the SegNet encoder decoder network \cite{badrinarayanan2015segnet}.
The encoder decoder network comprises of 4 layers for both encoder
and decoder, 7x7 convolutional layers and 64 features per layer. The
SegNet architecture is used to generate the per-pixel label for the
input sequences. There are total 12 different semantic class labels:
Sky, Building, Pole, Road Marking, Road, Pavement, Tree, Sign Symbol,
Fence, Vehicle, Pedestrian, and Bike. The SegNet architecture is used
here because of available trained models for urban segmentation tasks
and the ease of use. Note our framework is designed to incorporated
semantic information from any available methods, so this pre-trained
network can be replaced by any method that can generate a dense segmentation
labels on video frames.

\subsubsection{Gray-Scale Conversion}

Note the navigation system we used from \cite{chiu2016navigation}
focuses on visual feature extraction, tracking, and matching on gray-scale
video frames, not color images. Therefore, all the video data from
\cite{chiu2016navigation} used for experiments is grayscale. For
the purpose of evaluating the SegNet encoder decoder network \cite{badrinarayanan2015segnet},
we fine-tuned the network on the CamVid dataset \cite{brostow2009semantic}
for about 50 epochs with the images in the training sequence converted
to grayscale.

\subsubsection{Computation Improvement}

Note SegNet is not designed for real-time navigation applications.
To fulfill the computation requirements for navigation systems, we
improve the efficiency of the SegNet model while still maintaining
its accuracy by converting the model into a low rank approximation
of itself. The conversion is based on the method proposed by \cite{LowRank16}.
The configuration chosen for this approximation is shown in Table
\ref{tab:Kernels}. The segmentation time, which is the forward pass
performance time of the SegNet model, is therefore improved from 160
ms to 89 ms (almost 2x performance) to process one image on a single
Nvidia K40 GPU. Similar accuracy is maintained by fine-tuning this
low-rank approximation for approximately 4 epochs. For comparison,
we show the performance of the final low-rank approximation using
the same test sequences of the CamVid dataset against the original
pre-trained SegNet model. This comparison is shown in Table \ref{tab:performace_segnet}. 

\begin{table}
\caption{The table below shows the configuration that was used to generate
a low rank approximation of the SegNet architecture.}

\centering%
\begin{tabular}{|c|c|c|}
\hline 
\# of Kernels & Original  & Low Rank \tabularnewline
\hline 
\hline 
conv1\_1 & 64 & 8\tabularnewline
\hline 
conv1\_2 & 64 & 32\tabularnewline
\hline 
conv2\_2 & 128 & 32\tabularnewline
\hline 
conv3\_1 & 256 & 32\tabularnewline
\hline 
conv3\_2 & 256 & 64\tabularnewline
\hline 
conv3\_3 & 256 & 64\tabularnewline
\hline 
conv4\_1 & 512 & 64\tabularnewline
\hline 
conv4\_2 & 512 & 64\tabularnewline
\hline 
conv4\_3 & 512 & 128\tabularnewline
\hline 
conv5\_1 & 512 & 128\tabularnewline
\hline 
conv5\_2 & 512 & 128\tabularnewline
\hline 
conv5\_3 & 512 & 128\tabularnewline
\hline 
conv5\_3\_D & 512 & 128\tabularnewline
\hline 
conv5\_2\_D & 512 & 128\tabularnewline
\hline 
conv5\_1\_D & 512 & 128\tabularnewline
\hline 
conv4\_3\_D & 512 & 128\tabularnewline
\hline 
conv4\_2\_D & 512 & 64\tabularnewline
\hline 
conv4\_1\_D & 512 & 64\tabularnewline
\hline 
conv3\_3\_D & 256 & 64\tabularnewline
\hline 
conv3\_2\_D & 256 & 64\tabularnewline
\hline 
conv3\_1\_D & 256 & 32\tabularnewline
\hline 
conv2\_2\_D & 128 & 32\tabularnewline
\hline 
\end{tabular}\label{tab:Kernels}
\end{table}

\begin{table*}[t]
\caption{The table below shows the accuracy comparison between the original
pre-trained SegNet model and its low-rank approximation.}

\centering

\begin{tabular}{|c|c|c|c|c|c|c|c|c|c|c|c|c|c|}
\hline 
\begin{turn}{90}
\end{turn} & \begin{turn}{90}
Global Accuracy
\end{turn} & \begin{turn}{90}
Class Average
\end{turn} & \begin{turn}{90}
Mean IOU
\end{turn} & \begin{turn}{90}
Sky
\end{turn} & \begin{turn}{90}
Building
\end{turn} & \begin{turn}{90}
Pole
\end{turn} & \begin{turn}{90}
Road
\end{turn} & \begin{turn}{90}
Tree
\end{turn} & \begin{turn}{90}
Vehicle
\end{turn} & \begin{turn}{90}
Sign
\end{turn} & \begin{turn}{90}
Fence
\end{turn} & \begin{turn}{90}
Pedestrian
\end{turn} & \begin{turn}{90}
Bike
\end{turn}\tabularnewline
\hline 
\hline 
SegNet & 90.0207 & 72.097 & 60.15 & 90.43 & 93.72 & 56.72 & 95.88 & 93.04 & 89.66 & 56.24 & 61.54 & 98.52 & 86.23\tabularnewline
\hline 
Low Rank SegNet & 86.965 & 74.292 & 56.03 & 93.68 & 79.90 & 43.74 & 95.00 & 92.96 & 81.01 & 61.31 & 56.19 & 87.22 & 74.47\tabularnewline
\hline 
\end{tabular}\label{tab:performace_segnet}
\end{table*}

\subsection{Gated Factor graph}

\begin{figure}[t]
\centering\includegraphics{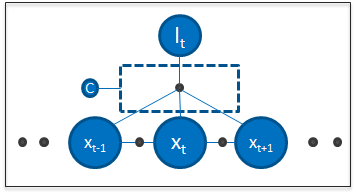}\caption{\label{fig:gatedfactor}The figure shows a portion of our gated factor
graph. The black dots represent the factors across navigation states
$x$ and landmark states $l$. The dotted lines represent the gated
approach and $c$ is the condition variable. }
\end{figure}

We proposed a new gated factor graph framework (Figure \ref{fig:gatedfactor}),
which is inspired by the work of \cite{minka2009gates}, to incorporate
semantic information on top of vision-based SLAM systems for vehicle
navigation applications. 

Factor graphs \cite{FactorGraph} are graphical models that are well
suited to modeling complex estimation problems, such as SLAM. A factor
graph is a bipartite graph model comprising two node types: \textit{factors
$f_{i}\in\mathcal{F}$} and \textit{state variables $\theta_{j}\in\Theta$.}
As shown in Figure \ref{fig:gatedfactor}, there are two kinds of
state variable nodes in our factor graph formulation for vision-based
SLAM systems: The navigation state nodes \textit{$X$} includes the
platform information (such as pose and velocity) at all given time
steps, while the landmark states \textit{$L$} encodes the estimated
3D position of external visual landmarks. Sensor measurements \textit{$z_{k}\in Z$}
are formulated into factor representations, depending on how a measurement
affects the appropriate state variables. For example, a GPS position
measurement only involves a navigation state \textit{$x$} at a single
time. A camera feature observation can involve both a navigation state
\textit{$x$} and a state of unknown 3D landmark position \textit{$l$}.
Estimating both navigation states and landmark positions simultaneously
is very popular in SLAM problem formulation, which is also known as
bundle adjustment \cite{bundle} in computer vision. 

The inference process of such a factor graph can be viewed as minimizing
the non-linear cost function as follows.

\begin{equation}
\sum_{k=1}^{K}||h_{k}(\Theta_{j_{k}})-\tilde{z_{k}}||_{\Sigma}^{2}
\end{equation}
where\textit{ $h(\Theta)$} is measurement function and and $||\cdot||_{\Sigma}^{2}$
is the Mahalanobis distance with covariance $\Sigma$. There are many
efficient solutions to solve this inference process for SLAM systems
using the factor graph representation. One popular solution is iSAM2
\cite{iSAM2}, which uses a Bayes tree data structure to keep all
past information and only updates variables influenced by each new
measurement. For the details on the factor graph representation and
its inference process for SLAM systems, we refer to \cite{FGGTSAM}.

Our new gated factor graph framework extends the factor graph representation
by modeling the semantic constraint as a gated factor (the dashed
lines in Figure \ref{fig:gatedfactor}) in the factor graph for the
inference process. As shown in Figure \ref{fig:gatedfactor}, a landmark
state $l_{t}$ is only added to the graph to participate the inference
process if the condition variable $c$ is $true$. Otherwise this
landmark is not used during the inference process in the vision-based
SLAM system. 

The value on the condition variable $c$ associated with a landmark
state is assigned based on the modes of semantic class labels from
all observations (2D visual features) on camera images for the same
3D visual landmark (Section \ref{subsec:Semantic-Segmentation}).
Note our SegNet architecture generates semantic class label for each
pixel on the image and there are total 12 different semantic class
labels. However, the same landmark may have different class labels
generated by SegNet for its observations across different video frames.
Therefore we accumulate the labels among 12 classes for all imaged
features correspondent to the same landmark, and decide the Boolean
value of the condition variable c for this landmark based on the final
counts among all 12 classes. If the landmark is classified as a ``valid''
semantic class based on the final counts, the condition variable $c$
becomes $true$.

The selection of valid semantic classes for gated factors is different
across three perspectives for the navigation system: the feature tracking
process, the geo-referenced map building process, and the navigation
system using pre-mapped landmarks. We describe how we define the ``valid''
semantic classes for each situation in the following subsections. 

\subsection{Visual Feature Tracking \label{subsec:Visual-Feature-Tracking}}

Our framework is able to directly improve real-time navigation performance
by enhancing the feature tracking process inside visual SLAM systems
(such as \cite{durrant2006simultaneous,fuentes2015visual,davison2007monoslam,davison2003real,ORBSLAM}).
To validate the influence of our approach in pure feature tracking
process, we apply our framework on top of the popular ORB-SLAM2 system
\cite{ORBSLAM}, which cannot incorporate GPS or pre-built geo-referenced
maps for navigation.

For each of the tracked features identified on the current video frame
from \cite{ORBSLAM}, our gated factor graph framework (Figure \ref{fig:gatedfactor})
makes inlier/outlier decision based on the modes of semantic class
labels from all 2D imaged positions tracked on past frames of the
same tracked feature. Visual features identified as non-static (such
as Pedestrian, Vehicle, Bike) or far-away classes (such as Sky) are
rejected, and will not contribute to the navigation solution in visual
SLAM systems.

\subsection{Geo-Referenced Map Building}

\begin{figure}[t]
\centering\includegraphics[scale=0.82]{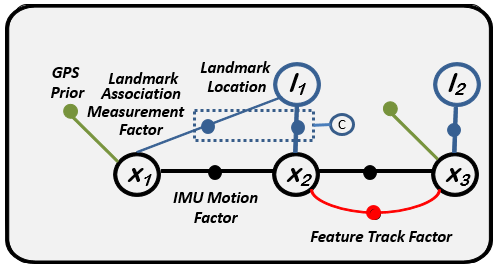}\caption{\label{fig:gaph_mapping}The figure above shows a section of our constructed
factor graph for the pre-mapping process in the system from \cite{chiu2016navigation}.
Factors are formed using measurements from GPS, IMU, and feature tracks.
Note factors formed from different kinds of sensor measurements are
shown as different colors. The black bubbles represent the state denoted
$x$ and the green bubbles represent a prior. The blue dots represent
measurements and the blue bubbles represent $l$ states. The dotted
lines represent our gated approach and $c$ is the condition variable. }
\end{figure}

For high-precision vision-based vehicle navigation systems (such as
\cite{chiu2016navigation,beall2014appearance}), our framework improves
both the mapping quality beforehand and the localization process using
pre-mapped visual landmarks. For demonstration, we implemented our
framework on top of the visual-inertial navigation system from \cite{chiu2016navigation}.
This system builds a high-quality, fully-optimized map of visual landmarks
beforehand using IMU, GPS, and one monocular camera. Our gated factor
graph for this geo-referenced mapping process is shown in Figure \ref{fig:gaph_mapping}.
Note there are GPS measurements (green factors in Figure \ref{fig:gaph_mapping})
used in the mapping process. 

We integrated our semantic segmentation module which classifies imaged
regions on input video frames (as shown in Figure \ref{fig:sem_seg})
into this system \cite{chiu2016navigation}. This system uses a landmark
matching module to construct the keyframe database, which is used
to match features across images to the key frame database. It also
has a visual odometry module which generates the features to track
across sequential video frames, and passes these features to both
the landmark matching module and the inference engine. Our semantic
segmentation module receives information about the tracked landmarks
from the landmark matching module, and generates the values for the
$c$ for the associated landmark constraints to the inference engine,
as shown in Figure \ref{fig:gaph_mapping}. Note only the 2D-3D visual
landmarks from selected semantic classes are preserved in our map.
The map of all 2D-3D semantic visual landmarks is then generated and
optimized. 

The semantic class label of the landmark is computed by using the
mode of the labels from all tracked 2D imaged features of the same
3D landmark. The value of the semantic categories is determined empirically.
The classes for visual landmarks that were found to be most useful
for mapping were Pole, Road Marking, Pavement, Sign Symbol, Tree,
Building, Fence. The semantic class labels that were rejected for
landmark selection were Sky, Pedestrian, Vehicle, Bike, Road. This
makes sense since the rejected classes of landmarks are associated
with non-static or far away objects. The road often does not add much
value visually since most of the extracted features from the road
are typically associated with shadows, which change over time. 

\subsection{High-Precision GPS-Denied Navigation}

\begin{figure}[t]
\centering\includegraphics[scale=0.78]{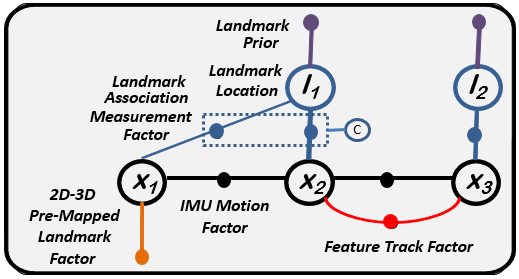}

\caption{\label{fig:final_graph}The figure above shows a section of our constructed
factor graph for the GPS-denied navigation system from \cite{chiu2016navigation}
using pre-mapped semantic visual landmarks. Factors are formed using
measurements from IMU, feature tracks, and pre-mapped visual landmark
observations. Note factors formed from different kinds of sensor measurements
are shown as different colors. The black bubbles represent the state
denoted $x$ and the green, orange and purple bubbles represent a
prior. The blue dots represent measurements and the blue bubbles represent
$l$ states. The dotted lines represent the gated approach and $c$
is the condition variable.}
\end{figure}

After the landmark database generated from the pipeline in Figure
\ref{fig:gaph_mapping}, GPS-denied navigation problem reduces to
estimate only the pose at the current time in the system from \cite{chiu2016navigation}.
The pre-built database is used by the landmark matching module to
match features tracked from the visual odometry module on new perceived
video frames, and provides absolute 2D-3D corrections through pre-mapped
visual landmarks in the inference engine. Note the tracked features
are also passed from visual odometry module to our semantic segmentation
module. This allows our semantic segmentation module to generate the
condition variable $c_{i}$values for all the constraints associated
with the tracked features during GPS-denied navigation, as shown in
Figure \ref{fig:final_graph}. Only visual features from selected
semantic classes are used in the inference engine.

\begin{figure}[t]
\centering\includegraphics[scale=0.19]{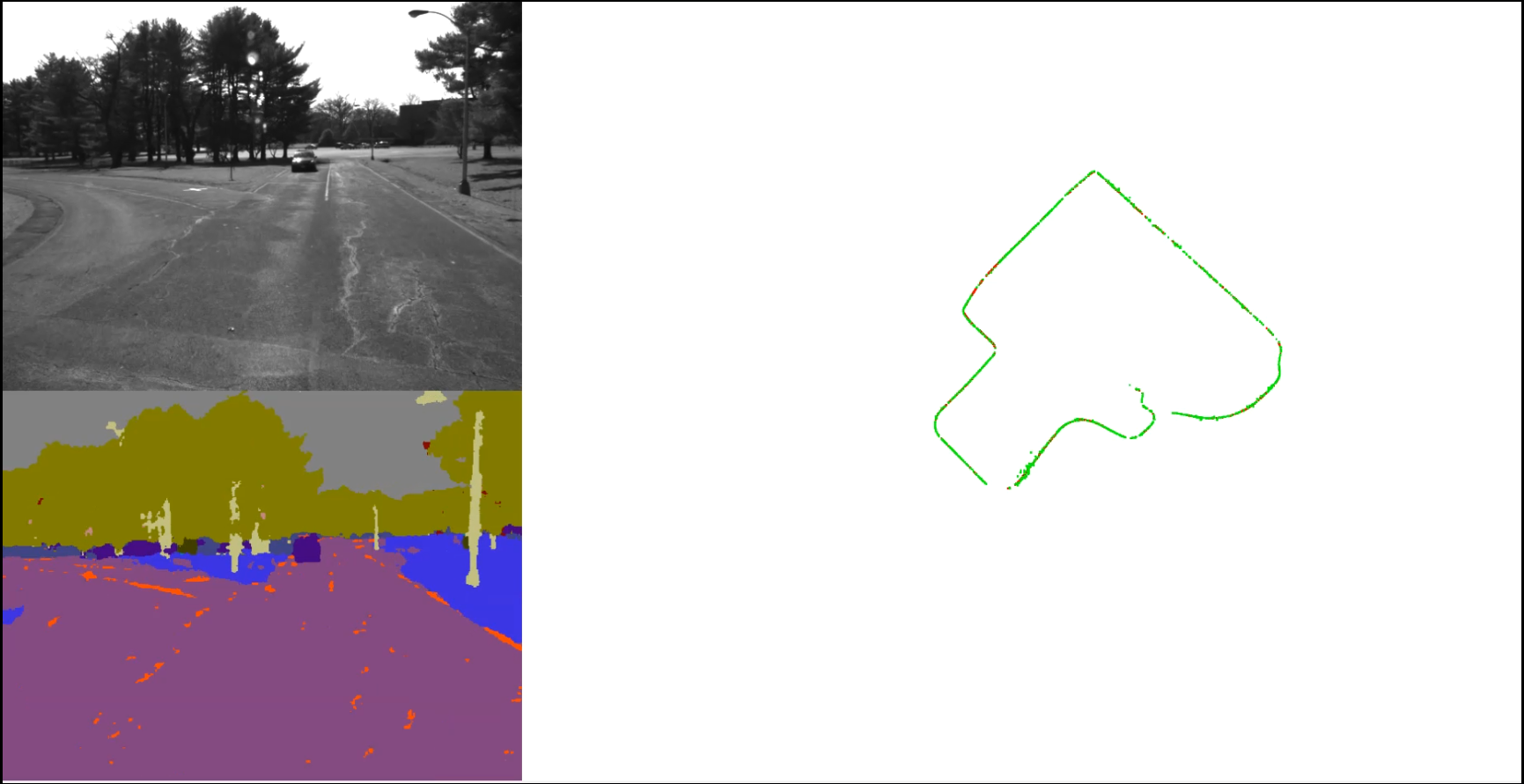}

\caption{\label{fig:system_visual}The figure above shows the visualization
of the system. The image on the top left is the current frame, and
the bottom left is the segmentation for that image. On the right,
the trajectory visualized in green is the estimated trajectory generated
by our system and the red trajectory shows the ground truth for this
sequence.}
\end{figure}

Our framework with the GPS-denied navigation system from \cite{chiu2016navigation}
is visualized in Figure \ref{fig:system_visual}. The pictures on
the left show the current image and the segmentation from our semantic
segmentation module for that image. The estimated trajectory generated
from our system is visualized in green, and the red trajectory shows
the ground truth for this sequence.

\section{Experimental Evaluation}

We validated our framework on top of two vision-based vehicle navigation
systems \cite{chiu2016navigation,ORBSLAM}, as described in Section
\ref{sec:Approach}. Based on the different characteristics of these
two systems, we conducted our experiments on various data sets to
demonstrate that our approach is able to utilize the semantic information
to improve the quality of the navigation solution from different perspectives.

\subsection{Feature Tracking for Navigation without Pre-Built Maps}

To demonstrate the improvement in feature tracking using our method,
we modified the ORB-SLAM2 system \cite{ORBSLAM} (Section \ref{subsec:Visual-Feature-Tracking})
which uses stereo camera input on the publicly available sequences
(sequences 00 - 10) from the KITTI localization benchmark \cite{geiger2012we}
without the use of pre-built maps. We keep the same configuration
for all sequences, and use the same configuration for both the baseline
performance and the results generated by combining our method with
\cite{ORBSLAM}. The parameters chosen for this configuration were
2000 features, 1.2 scale factor, and 8 levels in the scale pyramid.
The initial FAST threshold is 12, and the minimum FAST threshold is
7.

Note among these 11 KITTI sequences, there are 6 open-loop sequences
which cannot leverage the simultaneous on-the-fly mapping capability
from \cite{ORBSLAM} to correct drift when closing the loops during
navigation. For these open-loop sequences, the navigation accuracy
is purely decided by the visual feature tracking quality. As shown
in Table \ref{tab:LocationDrift}, our approach reduces approximately
0.1\% drift rate over 9.195 km distance by removing wrong tracked
features for these open-loop sequences. It also improves accuracy
for sequences with loop-closure optimization.

\begin{table*}
\caption{The table below shows our approach reduces the drift rate and mean
final location error of the ORB-SLAM2 system for 6 open-loop sequences
(open-loop), 5 sequences with loop closures (closed-loop), and all
11 sequences (overall) respectively. }
\centering%
\begin{tabular}{|c|c|c||c|c|}
\hline 
 & \multicolumn{2}{c||}{Location drift rate} & \multicolumn{2}{c|}{Mean Final Location Error}\tabularnewline
\hline 
Location drift rate & ORB-SLAM2 (\%) & ORSB-SLAM2+Ours (\%) & ORB-SLAM2 (m) & ORB-SLAM2 + Ours\tabularnewline
\hline 
\hline 
Open-Loop & 1.6103 & 1.5234 & 24.6778 & 23.3461\tabularnewline
\hline 
Closed-Loop & 0.5369 & 0.5292 & 13.7130 & 13.5164\tabularnewline
\hline 
Overall & 0.9862 & 0.9453 & 19.6938 & 18.878\tabularnewline
\hline 
\end{tabular}\label{tab:LocationDrift}
\end{table*}

\begin{figure}
\centering\includegraphics[scale=0.26]{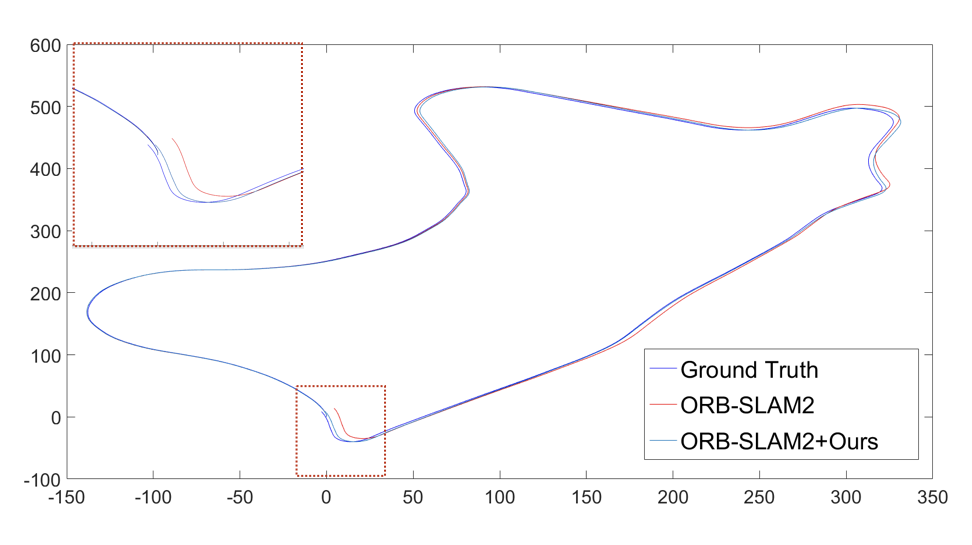}\caption{\label{fig:OpenLoop}The qualitative improvement (visualized in color)
on KITTI sequence 09, using our approach on top of ORB-SLAM2 system.
The x and y axes are in meters.}
\end{figure}

Figure \ref{fig:OpenLoop} shows the qualitative improvement in one
of the KITTI open-loop sequences (sequence 09). For this sequence,
there are moving vehicles in both directions during driving. Our method
avoids the use of tracked features on moving vehicles in ORB-SLAM2
system, and shows clear improvement over the standard ORB-SLAM2 performance
. For this sequence, the final error in location without semantic
selection is 12.6363 meters whereas the final location error is 3.9253
meters with out method. The total trajectory distance for this sequence
is 1.71 km. 

\subsection{Landmark Matching Using Semantic Selection\label{subsec:Improvements-in-Landmark}}

To demonstrate our framework for high-precision vehicle navigation
applications using pre-built maps, we applied our framework on top
of the system from \cite{chiu2016navigation}. We collected data across
seasons within same large-scale urban environments which include a
variety of buildings, highway driving, and lighting variations. The
vehicle we used for experiments incorporates a 100Hz MEMS Microstrain
GX4 IMU and one 20Hz front-facing monocular Point Grey camera. High-precision
differential GPS, which is also installed on the vehicle, was used
both for geo-referenced map construction and for ground truth generation
(fused with IMU) to evaluate our GPS-denied navigation system. All
three sensors are calibrated and triggered through hardware synchronization.
Note for testing this system, we are not aware of any publicly available
vehicle data that provides raw IMU and differential GPS measurements
with forward-facing camera inputs across season changes at the same
place.

\begin{figure*}[t]
\centering%
\begin{tabular}{ccc}
\subfloat{\includegraphics[height=4cm]{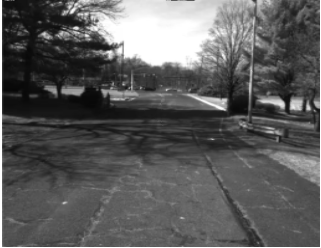}} & \subfloat{\includegraphics[height=4cm]{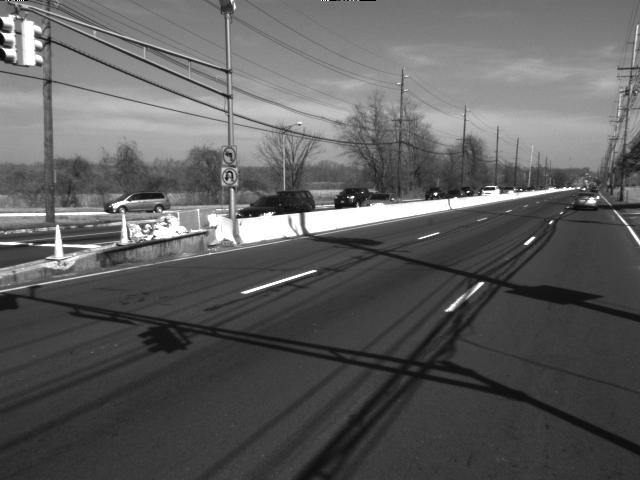}} & \subfloat{\includegraphics[height=4cm]{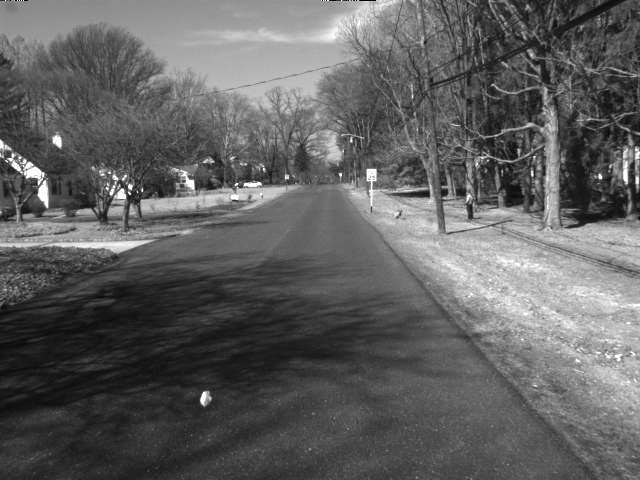}}\tabularnewline
 &  & \tabularnewline
\end{tabular}\caption{\label{fig:data_sequences}The figures above show representative images
in the sequences. The figure on the far left shows an image where
there are trees on both sides and no dynamic objects in the scene.
The middle images shows cars on both sides and is on the highway,
the right image shows houses on both sides and parked cars. These
3 types of environments are present in all the test sequences used
in the experiments.}
\end{figure*}

The total driving distance is 5.6 km and the total driving time is
around 10 minutes for each of the test sequences. Representative images
from the test sequences can be seen in Figure \ref{fig:data_sequences}.
Three sequences are used for our experiments, and they are collected
by driving in a clockwise loop along the same path around the campus.
The database for all experiments is constructed from one data sequence
collected in the morning on a partly cloudy day in winter.

\begin{table*}[t]
\caption{The table below shows the improvement in the landmark matching module
between using the semantic selection of the landmarks and traditional
landmark matching. The 3D mean/std error is from the camera re-sectioning
error given the pre-mapped 3D locations from the database sequence.}
\centering%
\begin{tabular}{|>{\centering}p{3cm}|c|c|}
\hline 
3D Position Accuracy & Without Semantic Selection (meter: mean/std deviation)  & With Semantic Selection (meter: mean/std deviation) \tabularnewline
\hline 
\hline 
Winter, Partly Cloudy Morning  & 1.871 / 5.745  & 1.457 / 3.756 \tabularnewline
\hline 
\hline 
Winter, Partly Cloudy Noon & 2.084 / 5.851  & 1.586 / 3.141 \tabularnewline
\hline 
\hline 
Spring, Sunny Morning  & 1.405 / 0.992  & 1.140 / 0.476 \tabularnewline
\hline 
\end{tabular}\label{tab:resectErr}
\end{table*}

\begin{table*}
\caption{The table below shows the 3D RMS error, 3D Median error, 3D 90 percentile
error in the visual-inertial navigation solution with/ without semantic
landmark selection.}

\centering%
\begin{tabular}{|>{\raggedright}p{1.8cm}|>{\centering}p{1.5cm}|>{\centering}p{1.5cm}|>{\centering}p{1.5cm}|>{\centering}p{1.5cm}|>{\centering}p{1.5cm}|>{\centering}p{1.5cm}|}
\cline{2-7} 
\multicolumn{1}{>{\raggedright}p{1.8cm}|}{} & \multicolumn{2}{c|}{3D RMS Error} & \multicolumn{2}{c|}{3D Median Error} & \multicolumn{2}{c|}{3D 90 percentile error}\tabularnewline
\cline{2-7} 
\multicolumn{1}{>{\raggedright}p{1.8cm}|}{} & without semantic (m) & with semantic selection (m) & without semantic (m) & with semantic selection (m) & without semantic (m) & with semantic selection (m)\tabularnewline
\hline 
Winter Partly Cloudy Noon & 0.5378 & 0.4207 & 0.3532 & 0.2856 & 0.7878 & 0.6109\tabularnewline
\hline 
Spring Sunny Morning & 1.1300 & 0.9640 & 0.7077 & 0.6227 & 1.8656 & 1.5476\tabularnewline
\hline 
\end{tabular}\label{tab:3Derrors}
\end{table*}

In this experiment, we evaluate the camera re-sectioning error and
the improvement that can be only gained by using semantic selection
for the pre-mapped visual landmarks. The results of this experiment
are shown in Table \ref{tab:resectErr}. As can be seen from Table
\ref{tab:resectErr}, the overall accuracy in the 3D error is reduced
by approximately 30 cm in all sequences and the standard deviation
is compressed in all the sequences which shows the improved robustness
of the system across seasonal variations using our approach for semantic
selection.

Note the camera resection error is computed using all the matched
2D-3D landmarks from the same single key frame stored in the database.
All landmarks with different 3D estimated uncertainty are treated
equally to compute the camera resection error. There are also outliers
in landmark matching results, which increase the error.

\subsection{Navigation Using Semantic Pre-Mapped Landmarks}

In this evaluation, we measure 3 error metrics to show the improvement
by using our semantic selection module for the entire visual inertial
navigation solution. Note there are some portions in the test sequence
where there are few or no pre-mapped landmarks available due to occlusion
or scene changes. However, the navigation system from \cite{chiu2016navigation}
continuously estimates 3D global pose by tightly fusing IMU data,
local tracked features from a monocular camera, and global landmark
points from associations between new observed features and pre-mapped
visual landmarks. It treats each new observation of a pre-mapped visual
landmark as a single measurement instead of computing only one pose
measurement from all landmark observations (such as in Section \ref{subsec:Improvements-in-Landmark})
at a given time. This way tightly incorporates geo-referenced information
into landmark measurements, and is capable of propagating precise
3D global pose estimates for longer periods in GPS-denied setting,
which results lower error than pure landmark matching error in Section
\ref{subsec:Improvements-in-Landmark}. 

The 3D RMS error from our GPS-denied navigation solutions is computed
across the whole test sequence and is compared to the ground truth
generated by the solution using differential GPS and the IMU. We compare
our solution using semantic information to the solutions presented
by \cite{chiu2016navigation}. The first metric is the 3D root mean
square error for the trajectory with and without the semantic selection
module. As can be seen in Table \ref{tab:3Derrors}, there is 21\%
improvement in the 3D RMS error. Table \ref{tab:3Derrors} also shows
the 3D median error and the 90 percentile error. As can be seen, our
approach improves the 3D Median error of the navigation solution by
19.1\% and improves the 3D 90 percentile error by 22\%.

\section{Conclusions}

In this paper, we present our framework to improve GPS-denied vehicle
navigation accuracy in large-scale urban environments, using semantic
information associated with visual landmarks. Our framework utilizes
the semantic information to improve the quality of the navigation
solution from three perspectives: the feature tracking process, the
geo-referenced map building process, and the navigation system using
pre-mapped landmarks. In comparison to previous state of the art techniques,
we show an improvement of around $20\%$ accuracy which is significant
for the precise vehicle navigation applications using pre-mapped visual
landmarks.

Compared to other semantic localization and mapping efforts, we present
a simple and yet effective approach to both construct landmark databases
and to perform localization. This could be further augmented by incorporating
cues from other types of information and is not limited to semantic
segmentation. We are also not dependent on using a single technique
for segmentation and this can be replaced by other techniques as the
state of the art improves.

For future work, we plan to accumulate better maps by aggregating
physically close features using their semantic labels and enhance
the visual map construction by intelligently gathering data from multiple
collections. Since we are already able to separate visual features
that are associated with permanent objects, the accuracy would only
improve with multiple runs since we can aggregate different information
to update the label associated with the landmark. We also plan on
experimenting with deep learned features in our system, that can be
trained to extract both a visual descriptor and a category label for
the landmark.

\bibliographystyle{IEEEtran}
\bibliography{IEEEabrv,IEEEexample,ICRA2017}

\begin{thebibliography}{10}
\providecommand{\url}[1]{#1}
\csname url@samestyle\endcsname
\providecommand{\newblock}{\relax}
\providecommand{\bibinfo}[2]{#2}
\providecommand{\BIBentrySTDinterwordspacing}{\spaceskip=0pt\relax}
\providecommand{\BIBentryALTinterwordstretchfactor}{4}
\providecommand{\BIBentryALTinterwordspacing}{\spaceskip=\fontdimen2\font plus
\BIBentryALTinterwordstretchfactor\fontdimen3\font minus
  \fontdimen4\font\relax}
\providecommand{\BIBforeignlanguage}[2]{{%
\expandafter\ifx\csname l@#1\endcsname\relax
\typeout{** WARNING: IEEEtran.bst: No hyphenation pattern has been}%
\typeout{** loaded for the language `#1'. Using the pattern for}%
\typeout{** the default language instead.}%
\else
\language=\csname l@#1\endcsname
\fi
#2}}
\providecommand{\BIBdecl}{\relax}
\BIBdecl

\bibitem{guizzo2011google}
E.~Guizzo, ``How google's self-driving car works,'' \emph{IEEE Spectrum Online,
  October}, vol.~18, 2011.

\bibitem{levinson2010robust}
J.~Levinson and S.~Thrun, ``Robust vehicle localization in urban environments
  using probabilistic maps,'' in \emph{Robotics and Automation (ICRA), 2010
  IEEE International Conference on}.\hskip 1em plus 0.5em minus 0.4em\relax
  IEEE, 2010, pp. 4372--4378.

\bibitem{badrinarayanan2015segnet}
V.~Badrinarayanan, A.~Handa, and R.~Cipolla, ``Segnet: A deep convolutional
  encoder-decoder architecture for robust semantic pixel-wise labelling,''
  \emph{arXiv preprint arXiv:1505.07293}, 2015.

\bibitem{cadena2016simultaneous}
C.~Cadena, L.~Carlone, H.~Carrillo, Y.~Latif, D.~Scaramuzza, J.~Neira, I.~D.
  Reid, and J.~J. Leonard, ``Simultaneous localization and mapping: Present,
  future, and the robust-perception age,'' \emph{arXiv preprint
  arXiv:1606.05830}, 2016.

\bibitem{choudhary2014slam}
S.~Choudhary, A.~J. Trevor, H.~I. Christensen, and F.~Dellaert, ``Slam with
  object discovery, modeling and mapping,'' in \emph{IROS 2014}, 2014.

\bibitem{chiu2016navigation}
H.-P. Chiu, M.~Sizintsev, X.~S. Zhou, P.~Miller, S.~Samarsekera, and R.~T.
  Kumar, ``Sub-meter vehicle navigation using efficient pre-mapped visual
  landmarks,'' in \emph{International Conference on Intelligent Transport
  Systems}, 2016.

\bibitem{ORBSLAM}
R.~Mur-Artal, J.~Montiel, and J.~Tard{\'o}s, ``Orb-slam: A versatile and
  accurate monocular slam system,'' \emph{IEEE transactions on robotics},
  vol.~31, no.~5, pp. 1147--1163, 2015.

\bibitem{durrant2006simultaneous}
H.~Durrant-Whyte and T.~Bailey, ``Simultaneous localization and mapping: part
  i,'' \emph{IEEE robotics \& automation magazine}, vol.~13, no.~2, pp.
  99--110, 2006.

\bibitem{fuentes2015visual}
J.~Fuentes-Pacheco, J.~Ruiz-Ascencio, and J.~M. Rend{\'o}n-Mancha, ``Visual
  simultaneous localization and mapping: a survey,'' \emph{Artificial
  Intelligence Review}, vol.~43, no.~1, pp. 55--81, 2015.

\bibitem{davison2007monoslam}
A.~J. Davison, I.~D. Reid, N.~D. Molton, and O.~Stasse, ``Monoslam: Real-time
  single camera slam,'' \emph{IEEE transactions on pattern analysis and machine
  intelligence}, vol.~29, no.~6, pp. 1052--1067, 2007.

\bibitem{davison2003real}
A.~J. Davison, ``Real-time simultaneous localisation and mapping with a single
  camera,'' in \emph{Computer Vision, 2003. Proceedings. Ninth IEEE
  International Conference on}.\hskip 1em plus 0.5em minus 0.4em\relax IEEE,
  2003, pp. 1403--1410.

\bibitem{chiu2013robust}
H.-P. Chiu, S.~Williams, F.~Dellaert, S.~Samarasekera, and R.~Kumar, ``Robust
  vision-aided navigation using sliding-window factor graphs,'' in
  \emph{Robotics and Automation (ICRA), 2013 IEEE International Conference
  on}.\hskip 1em plus 0.5em minus 0.4em\relax IEEE, 2013, pp. 46--53.

\bibitem{leutenegger2015keyframe}
S.~Leutenegger, S.~Lynen, M.~Bosse, R.~Siegwart, and P.~Furgale,
  ``Keyframe-based visual--inertial odometry using nonlinear optimization,''
  \emph{The International Journal of Robotics Research}, vol.~34, no.~3, pp.
  314--334, 2015.

\bibitem{reddy2015dynamic}
N.~D. Reddy, P.~Singhal, V.~Chari, and K.~M. Krishna, ``Dynamic body vslam with
  semantic constraints,'' in \emph{Intelligent Robots and Systems (IROS), 2015
  IEEE/RSJ International Conference on}.\hskip 1em plus 0.5em minus 0.4em\relax
  IEEE, 2015, pp. 1897--1904.

\bibitem{kundu2014joint}
A.~Kundu, Y.~Li, F.~Dellaert, F.~Li, and J.~M. Rehg, ``Joint semantic
  segmentation and 3d reconstruction from monocular video,'' in \emph{European
  Conference on Computer Vision}.\hskip 1em plus 0.5em minus 0.4em\relax
  Springer, 2014, pp. 703--718.

\bibitem{beall2014appearance}
C.~Beall and F.~Dellaert, ``Appearance-based localization across seasons in a
  metric map,'' \emph{6th PPNIV, Chicago, USA}, 2014.

\bibitem{alcantarilla2016street}
P.~F. Alcantarilla, S.~Stent, G.~Ros, R.~Arroyo, and R.~Gherardi, ``Street-view
  change detection with deconvolutional networks,'' in \emph{Robotics: Science
  and Systems}, 2016.

\bibitem{choudhary2015information}
S.~Choudhary, V.~Indelman, H.~I. Christensen, and F.~Dellaert,
  ``Information-based reduced landmark slam,'' in \emph{2015 IEEE International
  Conference on Robotics and Automation (ICRA)}.\hskip 1em plus 0.5em minus
  0.4em\relax IEEE, 2015, pp. 4620--4627.

\bibitem{biswas2014episodic}
J.~Biswas and M.~Veloso, ``Episodic non-markov localization: Reasoning about
  short-term and long-term features,'' in \emph{2014 IEEE International
  Conference on Robotics and Automation (ICRA)}.\hskip 1em plus 0.5em minus
  0.4em\relax IEEE, 2014, pp. 3969--3974.

\bibitem{meyer2010temporary}
D.~Meyer-Delius, J.~Hess, G.~Grisetti, and W.~Burgard, ``Temporary maps for
  robust localization in semi-static environments,'' in \emph{Intelligent
  Robots and Systems (IROS), 2010 IEEE/RSJ International Conference on}.\hskip
  1em plus 0.5em minus 0.4em\relax IEEE, 2010, pp. 5750--5755.

\bibitem{IV13}
H.~Lategahn, M.~Schreiber, J.~Ziegler, and C.~Stiller, ``Urban localization
  with camera and inertial measurement unit,'' in \emph{2013 IEEE Intelligent
  Vehicles Symposium (IV)}.\hskip 1em plus 0.5em minus 0.4em\relax IEEE, 2013.

\bibitem{brostow2009semantic}
G.~J. Brostow, J.~Fauqueur, and R.~Cipolla, ``Semantic object classes in video:
  A high-definition ground truth database,'' \emph{Pattern Recognition
  Letters}, vol.~30, no.~2, pp. 88--97, 2009.

\bibitem{LowRank16}
C.~Tai, T.~Xiao, Y.~Zhang, X.~Wang, and E.~Weinan, ``Convolutional neural
  networks with low-rank regularization,'' in \emph{2016 International
  Conference on Learning Representations (ICLR)}, 2016.

\bibitem{minka2009gates}
T.~Minka and J.~Winn, ``Gates,'' in \emph{Advances in Neural Information
  Processing Systems}, 2009, pp. 1073--1080.

\bibitem{FactorGraph}
F.~Kschischang, B.~Fey, and H.~Loeliger, ``Factor graphs and the sum-product
  algorithm,'' \emph{IEEE Trans. Information Theory}, vol.~47, no.~2, 2001.

\bibitem{bundle}
B.~Triggs, P.~F. McLauchlan, R.~I. Hartley, and A.~W. Fitzgibbon, ``Bundle
  adjustment - a modern synthesis,'' \emph{Lecture Notes in Computer Science},
  vol. 1883, pp. 298--375, 2000.

\bibitem{iSAM2}
M.~Kaess, H.~Johannsson, R.~Roberts, V.~Ila, J.~Leonard, and F.~Dellaert,
  ``isam2: Incremental smoothing and mapping using the bayes tree,''
  \emph{Intl. J. of Robotics Research}, vol.~31, pp. 217--236, 2012.

\bibitem{FGGTSAM}
F.~Dellaert, ``Factor graphs and gtsam: A hands-on introduction,'' in
  \emph{Georgia Institute of Technology}, 2012.

\bibitem{geiger2012we}
A.~Geiger, P.~Lenz, and R.~Urtasun, ``Are we ready for autonomous driving? the
  kitti vision benchmark suite,'' in \emph{Computer Vision and Pattern
  Recognition (CVPR), 2012 IEEE Conference on}.\hskip 1em plus 0.5em minus
  0.4em\relax IEEE, 2012, pp. 3354--3361.

\end{thebibliography}

\end{document}